%% file: main.tex
\documentclass[preprint,12pt,authoryear]{elsarticle}

\usepackage{amssymb}
\usepackage{amsmath}
\usepackage{textcomp}
\usepackage{gensymb}
\usepackage{lineno}
\usepackage{booktabs}
\usepackage{geometry}
\usepackage{algorithm}
\usepackage{algorithmicx}
\usepackage{algpseudocode}
\usepackage[utf8]{inputenc}
\usepackage{siunitx}
\usepackage{placeins}
\journal{Remote Sensing of Environment}

\begin{document}

\begin{frontmatter}

\title{Manual Labelling Artificially Inflates Deep Learning-Based Segmentation Performance on RGB Images of Closed Canopy: Validation Using TLS} %

\author[1]{M.J. Allen}
\author[1]{H.J.F Owen}
\author[2,3]{S.W.D. Grieve}
\author[1]{E.R. Lines}

\affiliation[1]{organization={Department of Geography, University of Cambridge},%
            city={Cambridge},
            postcode={CB2 3EN}, 
            country={UK}}

\affiliation[2]{organization={School of Geography, Queen Mary University of London},%
            city={London},
            postcode={E1 4NS}, 
            country={UK}}

\affiliation[3]{organization={Digital Environment Research Institute, Queen Mary University of London},%
            addressline={}, 
            city={London},
            postcode={E1 1HH}, 
            country={UK}}

\begin{abstract}
Monitoring forest dynamics at an individual tree scale is essential for accurately assessing ecosystem responses to climate change, yet traditional methods relying on field-based forest inventories are labor-intensive and limited in spatial coverage. Advances in remote sensing using drone-acquired RGB imagery combined with deep learning models have promised precise individual tree crown (ITC) segmentation; however, existing methods are frequently validated against human-annotated images, lacking rigorous independent ground truth. In this study, we generate high-fidelity validation labels from co-located Terrestrial Laser Scanning (TLS) data for drone imagery of mixed unmanaged boreal and Mediterranean forests. We evaluate the performance of two widely used deep learning ITC segmentation models - DeepForest (RetinaNet) and Detectree2 (Mask R-CNN) - on these data, and compare to performance on further Mediterranean forest data labelled manually. When validated against TLS-derived ground truth from Mediterranean forests, model performance decreased significantly compared to assessment based on hand-labelled from an ecologically similar site (AP50: 0.094 vs. 0.670). Restricting evaluation to only canopy trees shrank this gap considerably (Canopy AP50: 0.365), although performance was still far lower than on similar hand-labelled data. Models also performed poorly on boreal forest data (AP50: 0.142), although again increasing when evaluated on canopy trees only (Canopy AP50: 0.308). Both models showed very poor localisation accuracy at stricter IoU thresholds, even when restricted to canopy trees (Max AP75: 0.051). Similar results have been observed in studies using aerial LiDAR data, suggesting fundamental limitations in aerial-based segmentation approaches in closed canopy forests.

\end{abstract}



\end{frontmatter}

\include{sec_intro.tex}

\include{sec_methods}

\include{sec_results}

\include{sec_discussion}

\newpage
\appendix
\include{sec_app}

\bibliography{ch2}
\end{document}

%% file: sec_intro.tex
\section{Introduction}
\label{sec:intro}
Accurately monitoring the current and changing state of forests is essential to predicting the impact of climate change on both forest dynamics, and the ecosystem services they provide. Historically, monitoring efforts have relied on field-based forest inventories which capture forest dynamics at the individual tree level \citep{frejaville_inferring_2019}. This granularity is vital because the size, age and species of individuals, as well individual interactions, determine how both trees and whole forests respond to climate-induced stressors \citep{chen_strategies_2022,fernandez-de-una_role_2023,teskey_responses_2014,coomes_wood_2014}. Yet, the labor-intensive nature of these methods poses significant challenges, limiting their scope and effectiveness in covering expansive \citep{scott_forest_2002} or inaccessible forest areas \citep{zeng_national_2015}.  In contrast, satellite-based approaches offer a more scalable solution, but are limited by spatial resolution, often resulting in findings based on large-scale spectral averages rather than individual tree data - such as approaches using data from the Sentinel \citep{lastovicka_sentinel2_2020, allen_large_2024} or Landsat \citep{zhang_tracking_2021} satellite constellations. 

The increased availability of lower-cost and accessible airborne imagery from airplanes and drones \citep{ouaknine_openforest_2023,troles_bamforests_2024} combined with deep learning has revolutionized remote sensing-based forest monitoring by enabling large-scale, high-resolution observations of individual trees \citep{allen_lowcost_2024,karthigesu_improving_2024}. Often, a component of these approaches is the detection and delineation of canopy tree crowns in aerial imagery, for example to assess tree size distribution (crucial to understand forest carbon sequestration; \cite{stephenson_rate_2014}) or canopy health \citep{allen_lowcost_2024,sandric_trees_2022}. Recent years have seen a surge in studies using deep learning to precisely delineate individual trees from airborne imagery \citep{weinstein_individual_2019,ball_accurate_2023,allen_lowcost_2024,sandric_trees_2022,sani-mohammed_instance_2022,hao_automated_2021}, showcasing the potential for significant advancements in forest monitoring. However, a common theme among these studies is the lack of verification using independent ground truth data \citep{lines_ai_2022,yang_detecting_2022,sandric_trees_2022,hao_automated_2021}, instead typically relying on crown delineations created by hand from visual assessment of aerial images to create test, train and validation datasets. This introduces unquantified uncertainty into downstream tasks using individually segmented imagery. Rigorous testing of methods using independent data are crucial for reliable deployment. Recent work on the reliability of tree segmentation from aerial LiDAR data has shown it may be inaccurate for segmenting all but the tallest canopy crowns~\citep{cao_benchmarking_2023}, demonstrating the limitations of three dimensional data; and highlighting the need for independent ground-truthed testing of two-dimensional crown delineation approaches. %

Segmentation of individual trees from imagery is a very active research area. Early efforts, such as~\citet{onishi_explainable_2021}, showed some accuracy applying non-machine learning (ML) methods to RGB data. More recently, deep learning-based developments have dominated ~\citep{diez_deep_2021}, with studies relying on manually delineated data for test, train and validation. A seminal work in this area, DeepForest~\citep{weinstein_individual_2019}, applied a single-stage object detector (bounding box delineation), RetinaNet~\citep{lin_focal_2018}, pretrained on labels derived from aerial LiDAR and finetuned on hand-drawn labels of Californian forests, achieving precision and recalls of 0.69 and 0.61 respectively. Subsequent studies have applied instance segmentation, delineating non-rectangular crown polygons from imagery. Such methods often internally identify individual trees as bounding boxes, and then delineate crowns within these boxes. Mask R-CNN \citep{he_mask_2017a} is a popular architecture for crown delineation, often evaluated considering predictions with an intersection-over-union (IoU) of more than 0.5 as correct, and has been used, for example, to segment tropical forest canopies in French Guiana and Malaysia \citep{ball_accurate_2023}, plantations in China \citep{hao_automated_2021} and dead trees in Scotland \citep{chiang_deep_2020}. Using an IoU threshold of 0.5, these studies report F1 scores of 0.64 \citep{ball_accurate_2023} and 0.91 \citep{hao_automated_2021}, and a mean-average-precision of 0.54 \citep{chiang_deep_2020} respectively, with all studies relying on manually labelled data. 

Beyond Mask R-CNN, other deep learning-based instance segmentation models have shown promise on manually labelled data. \citet{ji_satellite_2024} applied BlendMask~\citep{chen_blendmask_2020} to high-resolution satellite data of low diversity forests near Beijing, achieving an F1-score of 0.86 at an IoU threshold of 0.5. \citet{speckenwirth_treeseg_2024} investigated multiple instance segmentation models on multispectral aerial images of unmanaged forest in Germany, with their best model achieving an F1-score of 0.84. \citet{li_deep_2023} developed a multitask network to produce tree counts and canopy maps from aerial data in Denmark at a national scale, although this approach did not delineate individual crowns. \citet{pedley_detecting_2025} combined DeepLabV3~\citep{chen_rethinking_2017} and SAM~\citep{kirillov_segment_2023a} for instance segmentation of urban trees in New Zealand, achieving an F1 score of 0.934 at an unspecified IoU threshold. Notably, the use of model composition - prompting SAM with semantic segmentation masks from DeepLabV3 - did not require individually segmented data for training, only evaluation.

The rise in deep learning approaches has been accompanied by a rise in initiatives providing open access crown instance segmentation data from aerial imagery. Five such data initiatives are listed in Table~\ref{tab:other-datasets}. BAMFORESTS~\citep{troles_bamforests_2024} manually delineated crowns using human experts, with a random subset verification through in-situ field visits. \citet{cloutier_influence_2024} combined manual orthomosaic delineation with GNSS field surveys marking individual trunk locations, likely ensuring a high degree of accuracy in the total tree count. SiDroForest~\citep{vangeffen_sidroforest_2022} contains two subsets: 872 manually validated crowns and 19,342 crowns automatically extracted using watershed segmentation on Structure-from-Motion (SfM) derived canopy height models. Point clouds derived from SfM are generally considered lower-fidelity than aerial LiDAR, which in turn may be inaccurate for segmenting all but the largest canopy crowns~\citep{cao_benchmarking_2023}. \citet{jansen_deep_2023} relied solely on manual delineation in ArcGIS using orthomosaic imagery, without additional validation steps. OAM-TCD~\citep{veitch-michaelis_oamtcd_2024} employed a multi-stage approach where paid non-expert annotators performed initial delineation with model assistance, followed by an expert review of the resulting annotations. The annotation protocol conservatively instructed annotators to segment closed canopy only if individual trees could be unambiguously identified, otherwise labeling such areas as “closed canopy.”

\begin{table}[t]
\caption{Comparison of Aerial Crown Instance Segmentation Datasets.\label{tab:other-datasets}}
\label{tab:crown-datasets}
\begin{center}
\begin{tabular}{lcccc}
\toprule
Dataset & Location & Area (ha) & Resolution & Crowns \\
\midrule
\textbf{BAMFORESTS}$^1$ & Bavaria, Germany & 105 & 1.7\,cm & 27,160 \\
Cloutier et al. & Quebec, Canada & 44 & 2\,cm & 22,933 \\
\textbf{SiDroForest}$^2$ & Siberia & -- & 3\,cm & 20,214$^*$ \\
Jansen et al. & N.\ Australia & 7 & 2\,cm & 2,547 \\
\textbf{OAM-TCD}$^3$ & Global & -- & 10\,cm & 336,000$^{**}$ \\
\bottomrule
\end{tabular}
\end{center}
\begin{flushleft}
{\small
$^*$Includes 872 manually labeled and 19,342 automatically extracted crowns\\
$^{**}$Approximately 280,000 individual trees and 56,000 tree groups\\[2pt]
Note: Area not specified for SiDroForest and OAM-TCD datasets\\[4pt]
{\footnotesize
$^1$\citet{troles_bamforests_2024}\\
$^2$\citet{vangeffen_sidroforest_2022}\\
$^3$\citet{veitch-michaelis_oamtcd_2024}}}
\end{flushleft}
\end{table}

Terrestrial Laser Scanning (TLS) data may offer rigorous independent data for aerial crown segmentation benchmarking when co-located with imagery. TLS data comprises point clouds taken using ground-based LiDAR sensors, which are able to capture forest structure at the millimetre to centimetre scale. Crucially, the additional high resolution structural information present in these data enables highly accurate individual tree segmentation - even using automated methods - to accuracies of centimetres or better \citep{wilkes_tls2trees_2023, wielgosz_point2tree_2023, xiang_accurate_2023, xiang_automated_2024, wielgosz_segmentanytree_2024}. This precision, which introduces spatial errors comparable to only a few pixels in aerial imagery, makes TLS-derived crown delineations an ideal ground truth reference.

In this work we generate instance segmentation labels for aerial imagery directly from co-located individually segmented TLS, an ultra high-fidelity ground-based LiDAR. By superimposing segmented trees from these data on orthoimagery, we generate high quality test data for evaluating two common methods. We answer the following questions:

\begin{enumerate}
    \item How well do pretrained segmentation models perform on unmanaged forest canopies in Mediterranean and boreal forests, when validated against high-fidelity ground truth data from an independent instrument?
    \item Does the measured performance differ to that on a similar ecosystem that has been labelled by hand?
    \item Does performance vary according to tree height, as seen for aerial LiDAR?
\end{enumerate}

%% file: sec_methods.tex
\section{Methods}
\subsection{Data}
\subsubsection{Study Areas}
We collected both TLS data and drone imagery from 15 plots in boreal forests in the \textbf{Joensuu} region of Eastern Finland in summer 2022, and 19 plots from Mediterranean forests in the \textbf{Alto Tajo} Natural Park in central Spain in summer 2021. These plots are in the FUNDIV exploratory forest plot network \citep{ratcliffe_biodiversity_2017}. Plots are unmanaged and contain both coniferous and deciduous trees, with up to three dominant canopy species in Finland (\textit{Picea abies, Pinus sylvestris and Betula spp.}) and up to four in Spain (\textit{Pinus nigra, P. sylvestris, Quercus ilex and Quercus faginea}). We collected drone imagery without TLS from nine plots in \textbf{Almorox} \citep{allen_lowcost_2024, moreno-fernandez_interplay_2022}, central Spain, in summer 2019, in plots dominated by \textit{Pinus pinea}. Both Spanish datasets represent unmanaged Mediterranean forests with structure and diversity impacted by water limitation, while Finnish data represent mixed boreal forests. All data were collected during the growing season, with trees leaf-on in all plots. A breakdown of the number of crowns visible from the air in each plot is available in \ref{app:dataset-summary}. As drone flights covered larger areas than the field plots, we use the term `area' to describe a region within a site covered by a single drone flight and corresponding orthomosaic, and the term `plot' to describe the subset area where TLS data was collected.

\subsubsection{Data Acquisition \& Processing}
\paragraph{TLS Point Clouds}
One plot of 30$\times$30 m was located in each area of the Joensuu and Alto Tajo sites. Plots were scanned using a Riegl VZ 400i with a TLS pulse repetition rate of \SI{600}{\kilo\hertz}. Scan locations were spaced in a grid pattern with \SI{10}{\meter} spacing, with both upright and horizontal scans in each scan position, and additional scans outside plot boundaries. Trees were individually segmented from the point cloud data using TLS2Trees~\citep{wilkes_tls2trees_2023}, followed by extensive manual refinement in CloudCompare 2.13.
\paragraph{Drone Imagery}
Drone flights for all sites and area  were conducted using a DJI Mavic Mini drone. Data was gathered with all relevant permissions and location-specific regulations observed, including the use of rotor guards. Raw images of dimensions 4000$\times$2250px were obtained from nadir and oblique (55\textdegree$\,$below horizontal; \cite{nesbit_enhancing_2019}) view directions, with 95\% front and 80\% side overlap, from flights between 10-20 m above top canopy height. Detailed information on individual areas can be seen in~\ref{app:dataset-summary}. Orthomosaics were generated using Agisoft Metashape 2.1.1 without the use of ground control points, as when used we found that inaccurate altitude measurements caused ground plane orientations that differed substantially from those generated by the TLS. We provide detailed observations on generating orthomosaics for co-location with TLS in~\ref{app:ortho}.

\subsubsection{Data Labelling}
\paragraph{TLS-Derived}
Visible crowns in the orthomosaic data from the Joensuu and Alto Tajo sites were segmented using the co-located, individually segmented, TLS data. The portions of each crown visible from above were first delineated without overlap according to Algorithm~\ref{alg:tls_pipeline}. The resulting delineations were then aligned with their corresponding orthoimages while maintaining the relative spacing between individual crowns. Due to minor geometric distortions in the orthoimagery, fine-scale manual adjustments were applied to individual crowns to achieve good alignment across the entire image extent, occasionally resulting in minimal crown overlap.

\begin{algorithm}
\begin{algorithmic}[1]
\Require Individually segmented trees (TLS\_Segs) in a common CRS
\For{each tree in TLS\_Segs}
    \State PDAL - Create DEM for tree \Comment{Resolution: 0.02m; Window size: 1px}
    \State Update \texttt{minx}, \texttt{miny}, \texttt{maxx}, \texttt{maxy} \Comment{Minimum, maximum x, y coordinates across all trees}
\EndFor
\State Compute extent = (\texttt{minx}, \texttt{miny}, \texttt{maxx}, \texttt{maxy})
\State GDAL - build empty VRT spanning plot \Comment{extent = (\texttt{minx}, \texttt{miny}, \texttt{maxx}, \texttt{maxy}), resolution = DEM resolution}
\State Read VRT as array \texttt{all\_array}
\State \quad Band 0: DEM \Comment{DEM value: top of canopy height at each pixel}
\State \quad Band 1: Tallest tree @ pixel \Comment{Index of tree corresponding to DEM at each pixel}
\For{each segment in TLS\_Segs}
    \State GDAL - build VRT \Comment{Match extent and resolution of empty VRT to align pixels}
    \State Read VRT as array
    \State Update DEM and tallest tree in \texttt{all\_array}
\EndFor
\State Polygonise VRT Band 1 \Comment{(Tallest tree @ pixel)}
\State Filter polygons: Keep largest by area and remove internal geometries
\end{algorithmic}
\caption{Pipeline for producing crown footprints aligned with orthoimagery from individually segmented point cloud data.\label{alg:tls_pipeline}}
\end{algorithm}

\paragraph{Manual Labelling}
Visible portions of crowns in the orthomosaic data from the Almorox site were delineated by hand (see \citet{allen_lowcost_2024}) from the orthomosaic data. This was performed in QGIS 3.30.

\subsection{Crown Segmentation using Deep Learning}
\subsubsection{Models}
\label{sec:models}
We applied two popular pretrained deep learning models to our data, DeepForest~\citep{weinstein_individual_2019} and Detectree2~\citep{ball_accurate_2023}. The architecture of DeepForest is based on RetinaNet~\citep{lin_focal_2018} and Detectree2 on Mask R-CNN~\citep{he_mask_2017a}. We chose not to retrain either model on our data for two reasons. Firstly - this is the most common use case for these tools, and leads to the most widely applicable conclusions. Secondly - as each plot did not cover an entire orthomosaic, the degree of false negative sampling (where trees are visible but not labelled outside plot boundaries) in the data from the sites with TLS data (Joensuu, Alto Tajo) is high, which complicates training. Evaluation is unaffected, as predictions were clipped to the extent of the TLS plots.

Data were downsampled such that input tiles were of compatible size with the pretrained weights of DeepForest ($400\times400$ px) and Detectree2 ($1000\times1000$ px).

\subsubsection{Evaluation}\label{sec:modeleval}
Both models were evaluated an all three sites - two labelled using co-located TLS (Joensuu, Alto Tajo) and one labelled by hand (Almorox).

For both DeepForest and Detectree2 we gridsearched across tile size and a non-maximum suppression (NMS) IoU jointly to maximise performance. A relative overlap of 0.5 across tiles was used for both models. At each gridsearch point, we calculate AP50 and AP75. We also report the best F1-score across confidence thresholds at the gridsearch point with the best AP50. 

For the gridsearch points for each model with the highest AP50, we provide a full set of precision recall curves, and also show precision-recall curves for canopy trees only. We define a canopy tree as one with a maximum height greater than or equal to $75\%$ of the maximum height within the same plot. To calculate precision on canopy trees - which requires including false positive predictions over a certain height - we assign a height values to prediction when more than $50\%$ of its area is covered by a ground truth label. Since the ground truth labels have (next-to) no overlap by design, it is only possible for one height to be assigned per-prediction. Predictions that could not be assigned a height were discarded for calculations on canopy trees.

Detailed results for gridsearches are included in~\ref{app:gridsearch}. To ensure comparability, we assess both models on bounding box performance only - although Detectree2 is capable of producing non-rectangular delineations.

%% file: sec_results.tex
\section{Results}
\FloatBarrier
Performance metrics for both DeepForest (DF) and Detectree2 (DT) can be seen in Table~\ref{tab:metrics}. Detectree2 outperformed DeepForest on all metrics (AP50, AP75, F1) across all three sites with the sole exception of AP75 on the data from Joensuu (Finland) - although the scores for both models are likely too low  (DeepForest: 0.005; Detectree2: 0.002) to be significant. Notably, the performance gap was significant (Detectree2: 0.375; DeepForest: 0.036) when assessed at a strict IoU threshold of 0.75 on the manually delineated data.

Both models showed significantly lower performance overall on the TLS-labelled sites (Joensuu, Alto Tajo) compared to the manually labelled data (Almorox), which showed good performance (Best AP50: 0.670, AP75: 0.375, F1: 0.674). AP50 shrank from 0.385 (DeepForest) and 0.670 (Detectree2) to 0.05 (DeepForest) and 0.094 (Detectree2) when labels were generated using TLS (Alto Tajo) rather than labelled by hand (Almorox) in similar ecosystems. AP75 similarly shrank from 0.036 (DeepForest) and 0.375 (Detectree2) to 0.002 (DeepForest) and 0.011 (Detectree2). Maximum F1 scores at an IoU threshold of 50 shrank from 0.523 (DeepForest) and 0.674 (Detectree2) to 0.196 (DeepForest) and 0.227 (Detectree2).

The performance difference between manual and TLS data was significantly smaller when only considering canopy trees, at an IoU threshold of 0.5. AP50, for example, rose from 0.05 to 0.161 for DeepForest and 0.094 to 0.365 for Detectree2 on the TLS-labelled data from Alto Tajo - although still significantly lower than on the manually labelled data from a similar ecosystem (Best AP50: 0.670). Performance was still poor on both ecosystems when evaluated against the TLS data with a strict IoU threshold of 0.75 (Best AP75; Joensuu: 0.011, Alto Tajo: 0.051), regardless of whether models were assessed on canopy or all trees.

\begin{figure}
\centering
\includegraphics[width=15cm]{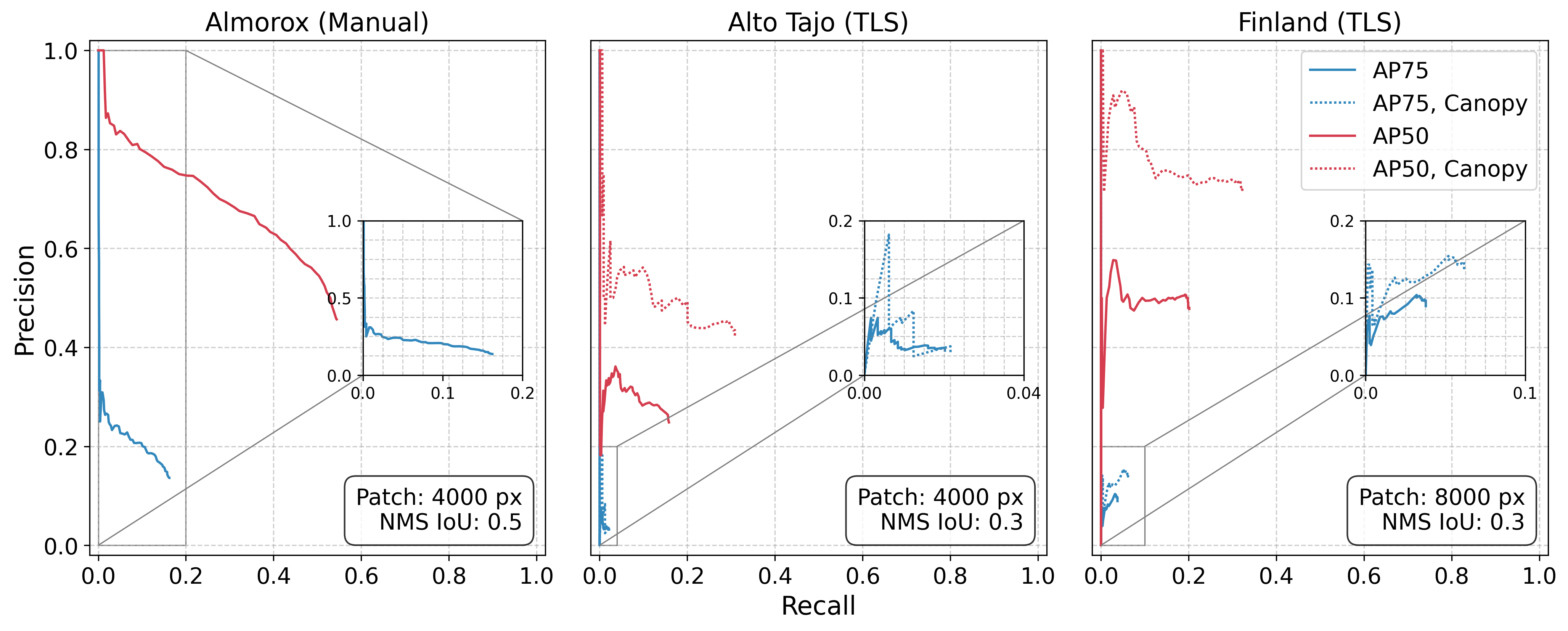}
\includegraphics[width=15cm]{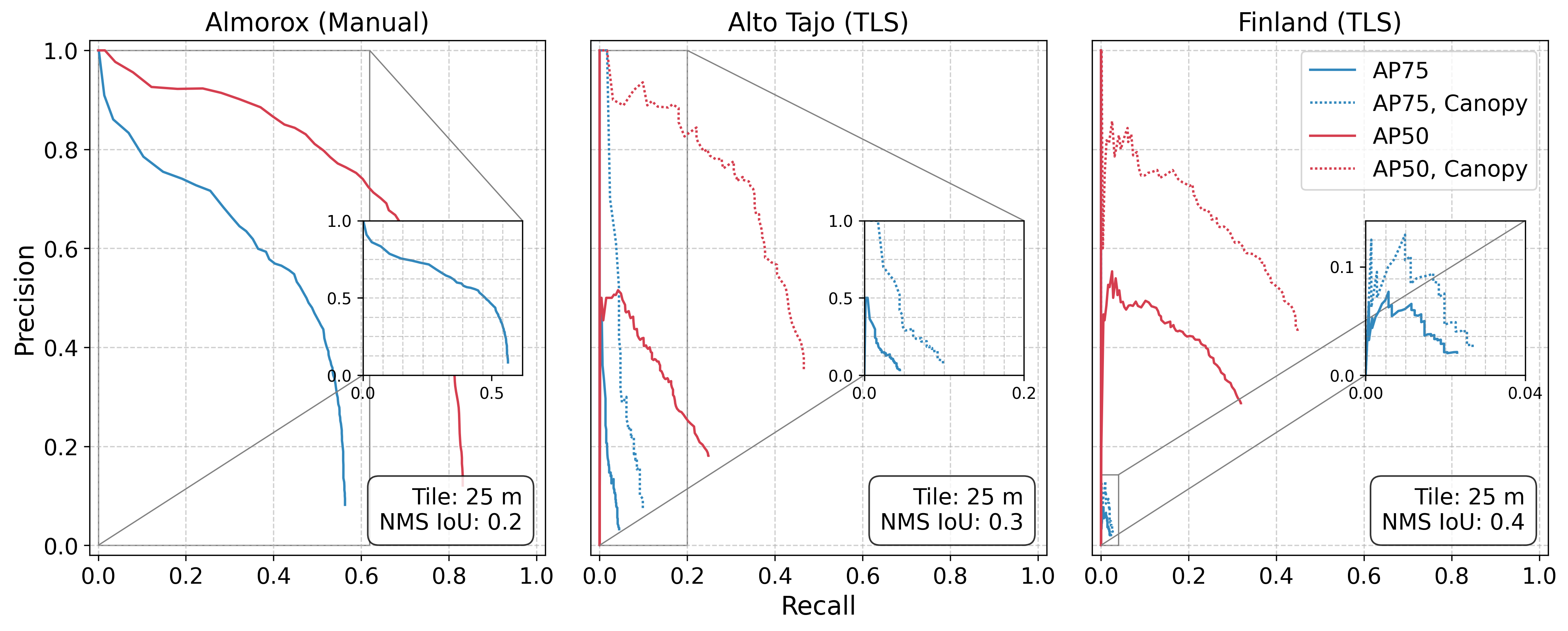}

\caption{Precision-recall curves for \textbf{(top)} Deepforest and \textbf{(bottom)} Detectree2 at the best hyperparameters for each site. Performance was assessed at IoU thresholds of 0.5 (red lines) and 0.75 (blue lines). Precision-recall curves obtained when scoring against canopy trees only are shown using dashed lines for both IoU thresholds.}\label{fig:pr_df}
\end{figure}

Precision-recall curves for all sites, for both Deepforest and Detectree2, can be seen in Figure~\ref{fig:pr_df}. Detectree2 (bottom row) outperformed Deepforest (top row) consistently. Both showed reasonable overall performance on manually labelled data (left column), although the detection ability of DeepForest dropped significantly when assessed at an IoU threshold of 0.75 (blue lines) versus 0.5 (red lines). Detectree2 showed good performance on the manually labelled data at both thresholds. Both models showed consistently higher performance on the TLS data for canopy (Deepforest/Detectree2 AP50 Alto Tajo: 0.161/0.365, Joensuu: 0.257/0.308) than overall (Deepforest/Detectree2 AP50 Alto Tajo: 0.094/0.050, Joensuu: 0.142/0.105), although neither showed good performance on canopy trees at an IoU threshold of 0.75 (Deepforest/Detectree2 AP75 Alto Tajo: 0.005/0.051, Joensuu: 0.011/0.004).

\begin{table}
  \caption{Best parameters and corresponding metrics for \textbf{DeepForest (DF)} and \textbf{Detectree2 (DT)}. A relative overlap of 0.5 between tiles was used throughout. Canopy (Can) trees are $\geq 75\%$ of the maximum height within each plot. NMS IoU refers to the minimum overlap for removal during post-processing, as opposed to the IoU correctness threshold used to calculate metrics. Reported F1-scores are maximised over confidence threshold, at an IoU threshold of 0.5.}
  \label{tab:metrics}
  \centering
  \begin{tabular}{ll@{\hspace{0.5em}}c@{\hspace{0.5em}}c@{\hspace{0.5em}}cc@{\hspace{0.5em}}cc@{\hspace{0.5em}}cc}
    \toprule
    & & & & \multicolumn{2}{c}{AP$_{50}$} & \multicolumn{2}{c}{AP$_{75}$} & \multicolumn{2}{c}{F1} \\
    \cmidrule(lr){5-6} \cmidrule(lr){7-8} \cmidrule(lr){9-10}
    Site & Mod & Tile Size & NMS IoU & All & Can & All & Can & All & Can \\
    \midrule
    Almorox & DF & 4000 px & 0.5 & 0.385 & -- & 0.036 & -- & 0.523 & -- \\
    (Manual) & DT & \SI{25}{\meter} & 0.2 & 0.670 & -- & 0.375 & -- & 0.674 & -- \\
    \midrule
    Alto Tajo & DF & 4000 px & 0.3 & 0.050 & 0.161 & 0.002 & 0.005 & 0.196 & 0.361 \\
    (TLS) & DT & \SI{25}{\meter} & 0.3 & 0.094 & 0.365 & 0.011 & 0.051 & 0.227 & 0.472 \\
    \midrule
    Joensuu & DF & 8000 px & 0.3 & 0.105 & 0.257 & 0.005 & 0.011 & 0.284 & 0.445 \\
    (TLS) & DT & \SI{25}{\meter} & 0.4 & 0.142 & 0.308 & 0.002 & 0.004 & 0.311 & 0.458 \\
    \bottomrule
  \end{tabular}
\end{table}

\begin{figure}
\centering
\includegraphics[width=15cm]{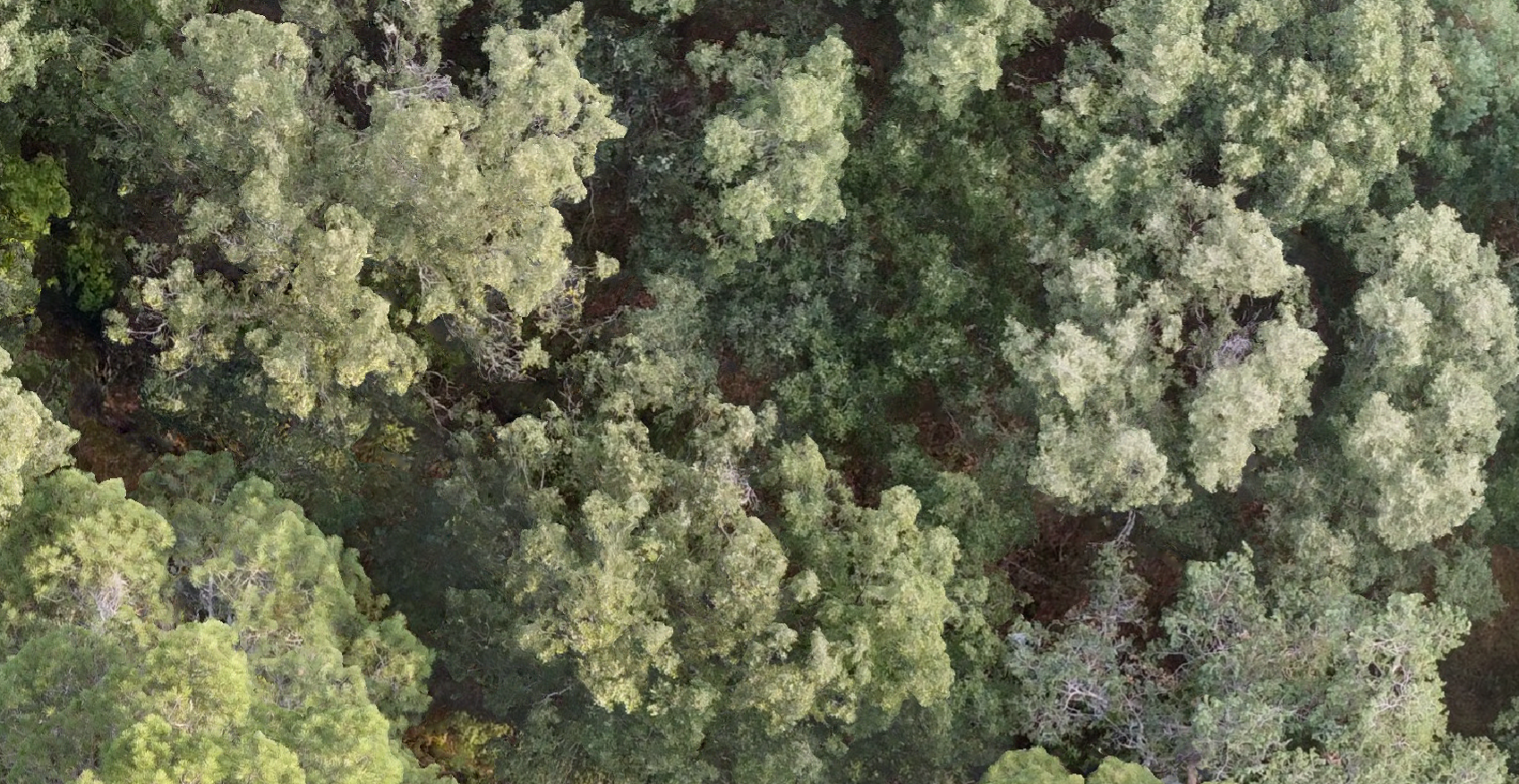}
\includegraphics[width=15cm]{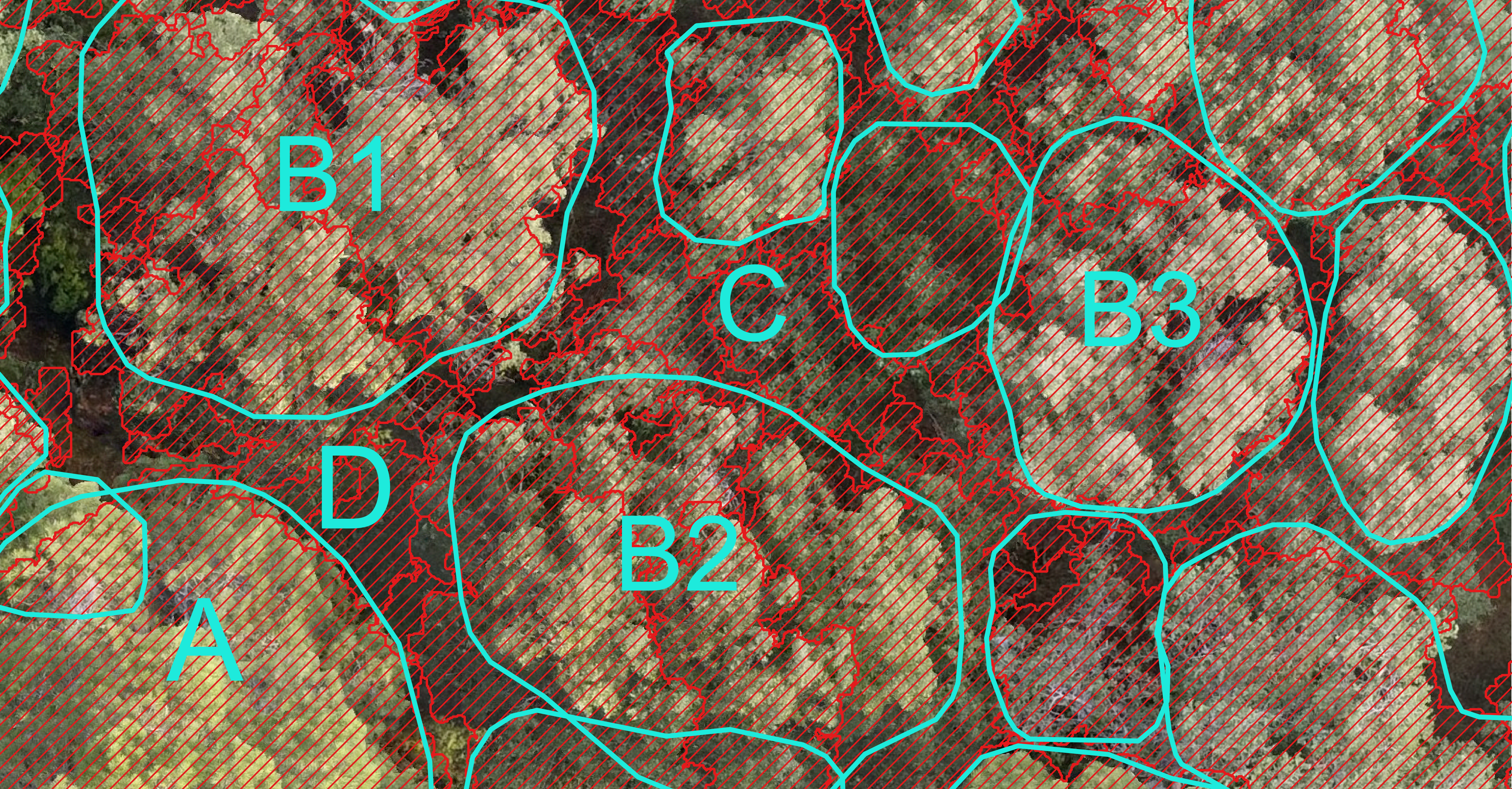}

\caption{\textbf{(Top)} Image data  with \textbf{(Bottom)} accompanying TLS-derived labels (Hashed, red) and Detectree2 predictions (Outlined, green) using the best hyperparameters. Several common sources of error are shown. \textbf{(A)} Large canopy tree segmented correctly. \textbf{(B1-3)} Canopy trees where the correct delineation is visually ambiguous - \textbf{(B1,2)} depict tightly grouped individual trees. \textbf{(B3)} is visually similar from above but is a single tree. \textbf{(C)} Sub-canopy trees that might be possible to predict but are generally ommitted during hand-labelling. \textbf{(D)} Sub-canopy trees that are nominally visible from above but practically invisible due to shadowing or orthomosaic artefacting.} \label{fig:pics}
\end{figure}

\begin{figure}
\centering
\includegraphics[width=4.9cm]{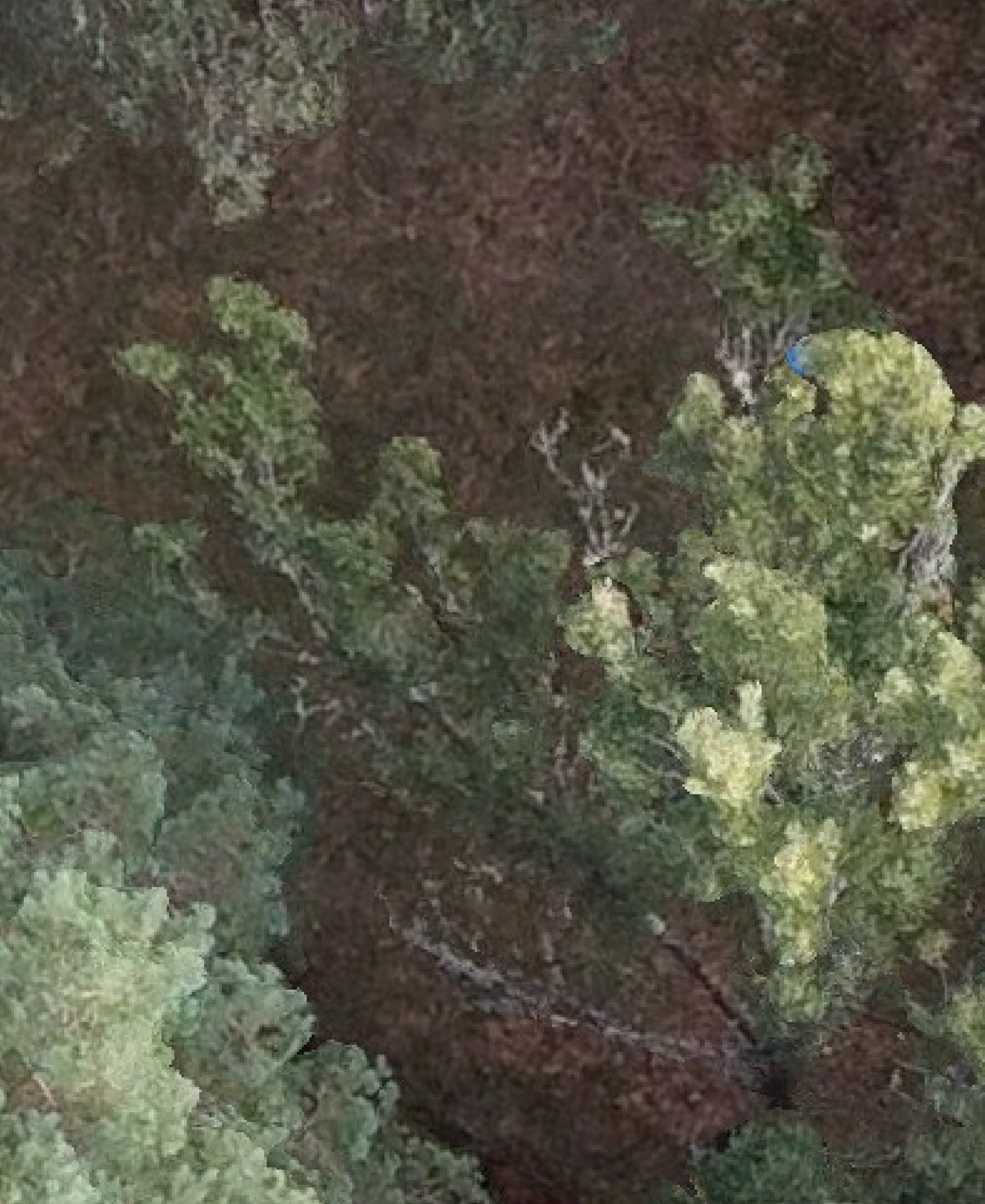}
\includegraphics[width=4.9cm]{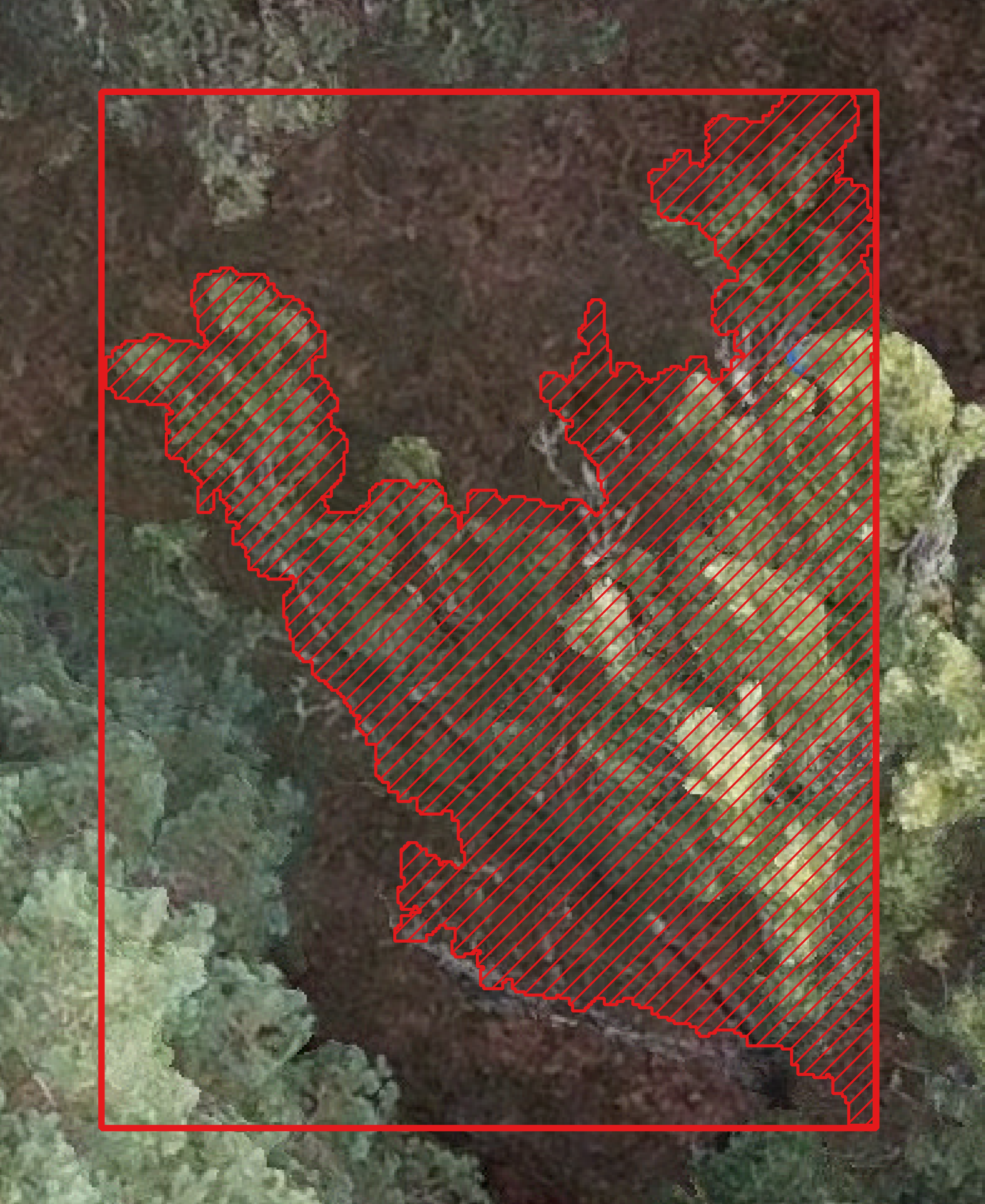}
\includegraphics[width=4.9cm]{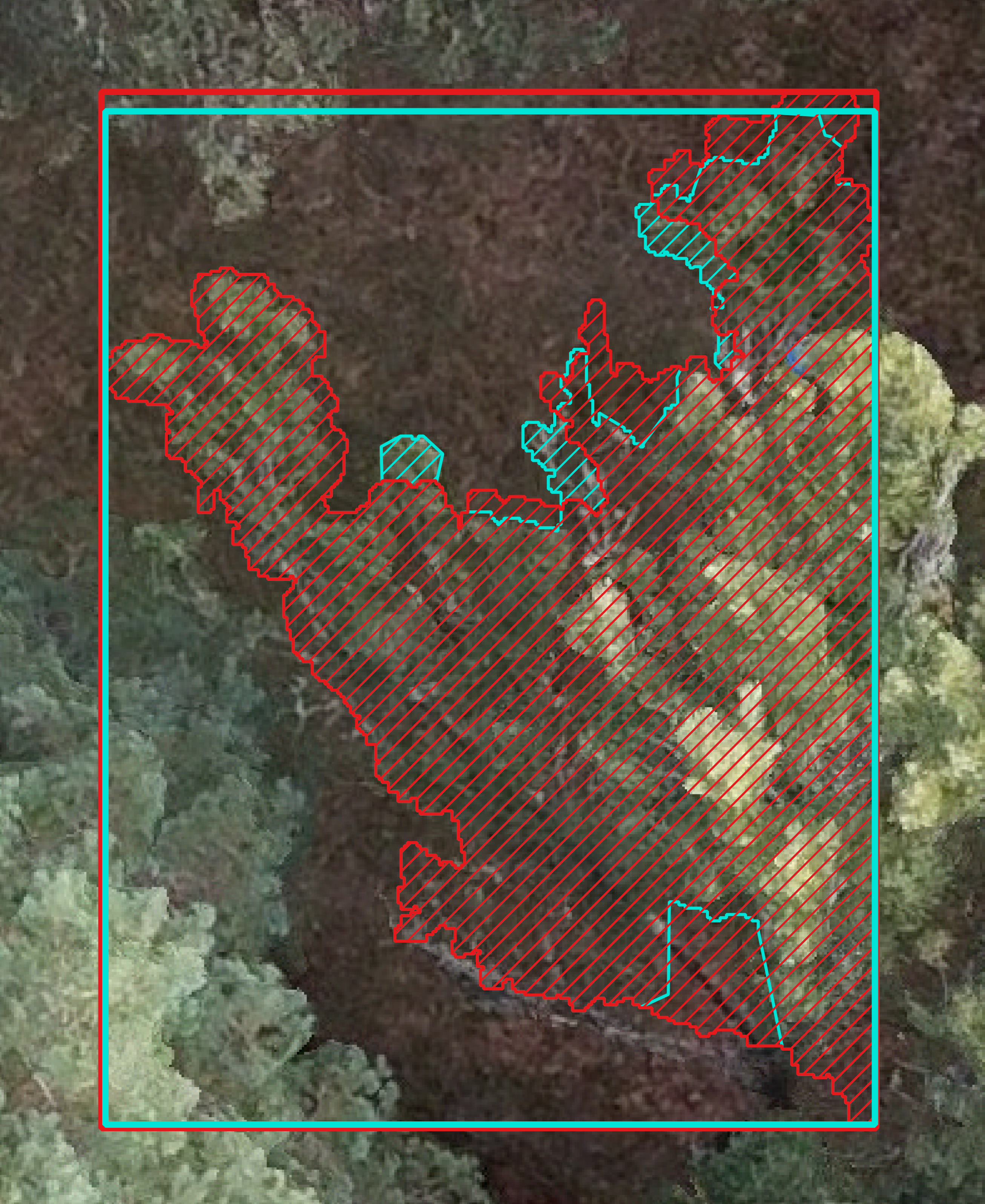}

\caption{Example of minor misalignment between TLS and image data. \textbf{(Left)} Raw imagery of a single tree. \textbf{(Middle)} Output polygon from Algorithm~\ref{alg:tls_pipeline}, with bounding box added for illustration. Ground truth polygons removed for surrounding trees to reduce visual clutter. Note cropping due to plot-edge effects. Predictions were also cropped to the extent of the plot after all other post-processing to eliminate the influence of plot-edge effects on evaluation. \textbf{(Right)}. Manually corrected polygon and bounding box (blue) overlaid on originals (red). The minor misalignment of individual branches causes a more pronounced shift in the full polygon than the bounding box. We evaluated segmentation using bounding boxes only.}\label{fig:alignment}
\end{figure}

\FloatBarrier

%% file: sec_discussion.tex
\section{Discussion}
\label{sec:discussion}

Our results show substantially lower performance for pretrained segmentation models on mixed unmanaged forest canopy when validated against high-fidelity ground truth data than previously reported in literature. Both study sites using TLS data showed poor AP50 values (0.105/0.142 and 0.050/0.094 for DeepForest/Detectree2 on Joensuu and Alto Tajo respectively; Table~\ref{tab:metrics}) and very poor AP75 scores (Joensuu: 0.005/0.002, Alto Tajo: 0.002/0.011). Even when maximising by adjusting confidence thresholds - which likely gives the best case scenario for model performance - the F1-scores were poor on both sites for both models (Joensuu: 0.284/0.311, Alto Tajo: 0.196/0.227). These metrics are markedly lower than those typically reported in crown delineation studies. \citet{ball_accurate_2023} reported an F1-score of 0.64 on tropical forest, \citet{hao_automated_2021} reported a F1-score of 0.85 on plantation forest in China and \citet{weinstein_individual_2019} reported an AP50 of 0.61 on open canopy forest in North America. Notably, all of these studies, along with others, report recalls almost exclusively higher than 0.5 \citep{ji_satellite_2024, sandric_trees_2022, speckenwirth_treeseg_2024, sun_yolov8e_2023, yang_detecting_2022, zhang_multispecies_2022}, which was not achieved in any case on the TLS data (Figure~\ref{fig:pr_df}). The performance of DeepForest was lower than that of Detectree2 throughout. This may stem from both differences in training data, with DeepForest being trained on open, single-story canopy data (temperate) and Detectree2 on closed canopy (albeit from a tropical ecosystem, unlike our evaluation data), or input resolution - with DeepForest using tiles of $400\times400$ px and Detectree2 $1000\times1000$ px (internally - input tiles are downsampled to match these sizes). For the remainder of this section, we discuss results from Detectree2 only, although the lighter computational footprint of DeepForest could lend itself to large open canopy areas.

The performance disparity between TLS-validated and manually labeled sites from similar ecosystems was notable, with AP50 increasing from 0.094 to 0.670, AP75 from 0.011 to 0.375 and maximum F1 score from 0.227 to 0.674 on the manually labeled Almorox site, despite being ecologically similar to Alto Tajo. While very minor misalignment between TLS-derived labels and orthoimagery might affect performance metrics, our use of metrics based on bounding boxes rather than non-rectangular delineations should be permissive (Figure~\ref{fig:alignment}) enough to mitigate this effect.

The performance difference between canopy ($\geq 75\%$ max plot height) and non-canopy trees offers some insight regarding this difference. Clear increases were observed for all metrics on both TLS sites (Joensuu: AP50 $0.142\rightarrow 0.308$, AP75 $0.002\rightarrow 0.004$, F1 $0.311\rightarrow 0.458$; Alto Tajo:  AP50 $0.094\rightarrow 0.365$, AP75 $0.011\rightarrow 0.051$, F1 $0.311\rightarrow 0.458$). Notably, the range of precisions achieved in Figure~\ref{fig:pr_df} (bottom row) for canopy trees on the TLS data (right two columns, dashed lines) are similar to the precisions achieved on the manually labelled data, at least when using an IoU threshold of 0.5. Previous work reports similar results on aerial LiDAR data - with crown segmentation only accurate for canopy trees~\citep{cao_benchmarking_2023}. Large canopy trees are visually obvious, and more likely to be included in manual training data as well as providing clear patterns for models to predict on.

Although the canopy precision is similar to the manually labeled data, the resulting average precision (AP50) is much lower because the recall is poorer. The phenomenon of poor detection rates in automated individual tree detection is often observed when using other aerial data sources such as LiDAR~\citep{cao_benchmarking_2023, kaartinen_international_2012}, and our results on validating RGB-based segmentation appear to confirm this pattern. We offer two hypotheses as to why this is the case. Firstly, many sub-canopy trees are not labelled at all in manually delineated training data. Often, these trees are missed due to being barely visible due to either being small or in shadow (See Figure~\ref{fig:pics}), even if they might theoretically be delineated from aerial images. Secondly - many trees are covered by labels, but incorrectly. Where there is significant intersection or overlap between crowns, visual separation is often ambiguous. We show an example of such a case in Figure~\ref{fig:pics} (B1-3). In Figure~\ref{fig:pr_df}, it can be seen that even though the precisions at an IoU threshold of 0.5 are similar between the manual labels and canopy trees from TLS-labelled sites, the precisions are far poorer when using a stricter threshold of 0.75. This pattern is not observable in the manually labelled data, where performance is good at both thresholds. This huge drop in performance points to poor localisation ability - the models are able to detect the presence or absence of trees, but show little ability to separate crowns precisely in closed canopy. Manually labelled data suffers from a bias towards obvious canopy breaks - which do not necessarily correspond to individual trees (Figure~\ref{fig:pics} B1-B3). While it is possible that training models at significantly higher per-pixel resolutions and using advanced strategies to merge tiled predictions \citep{allen_lowcost_2024} could solve this problem, it is not clear that the necessary structural information is observable from above. Decennially spaced benchmarking efforts for tree crown detection in aerial LiDAR data show only very small improvements in detection rates \citep{kaartinen_international_2012, cao_benchmarking_2023}, despite the explosion an machine learning-based methods over the same time frame - suggesting a fundamental challenge in detecting and delineating individual trees from aerial data, regardless of sensing technology. In some ecosystems, high species diversity can lead to higher variation in canopy reflectance, which might aid in crown separation~\citep{ball_accurate_2023}, but our results suggest that RGB-based segmentation is unlikely to be robust in forests.

These limitations have particular relevance for national-scale forest inventory or tree counting initiatives \citep{sun_counting_2022, li_deep_2023}. Individual tree mapping from RGB imagery alone may be infeasible in closed canopy forests, which are common in boreal, temperate and wet tropical ecosystems. While measurements of total canopy cover could be reliable, and crown delineation may be feasible in open canopy environments, attempts to count or delineate individual trees in closed canopy are likely to produce systematically inaccurate results. Given that similar observations have been made on segmentation using aerial LiDAR \citep{cao_benchmarking_2023}, this could be a fundamental limitation of the aerial perspective rather than a methodological shortcoming --- the necessary structural-spectral information simply cannot be captured from above. We caution that the use of individual tree data from aerial imagery without ground validation may be unreliable, and that the approach adopted by OAM-TCD~\cite{veitch-michaelis_oamtcd_2024}, with individual delineation only in open canopy, is more trustworthy in this case. Rapid national-scale forest inventories or other monitoring efforts seeking individual tree data may require alternative or complementary approaches to aerial sensing.

\subsection*{Author Contributions}
\textbf{Conceptualisation}: All authors. \textbf{Data Curation}: M. J. Allen, E. R. Lines, H. J. F. Owen. \textbf{Formal Analysis}: M. J. Allen, S. W. D. Grieve, E. R. Lines. \textbf{Investigation}: M. J. Allen, E. R. Lines. \textbf{Visualisation}: M. J. Allen. \textbf{Writing - original draft}: M. J. Allen. \textbf{Writing - review and editing}: S. W. D. Grieve, E. R. Lines, H. J. F. Owen. \textbf{Supervision}: S. W. D. Grieve, E. R. Lines.

\subsection*{Data \& Code Availability}
Data and code are undergoing final preparations and will be made available through Zenodo and Github respectively prior to publication. Intermediate results such as model predictions will be provided in addition to raw data.

\subsection*{Funding Sources}
M. J. Allen was supported by the UKRI Centre for Doctoral Training in Application of Artificial Intelligence to the study of Environmental Risks [EP/S022961/1]. E. R. Lines, S. W. D. Grieve and H. J. F. Owen were funded by a UKRI Future Leaders Fellowship awarded to E. R. Lines [MR/T019832/1].

%% file: sec_app.tex
\FloatBarrier
\section{Gridsearch Results}
\label{app:gridsearch}
\begin{figure}
\centering
\includegraphics[width=15cm]{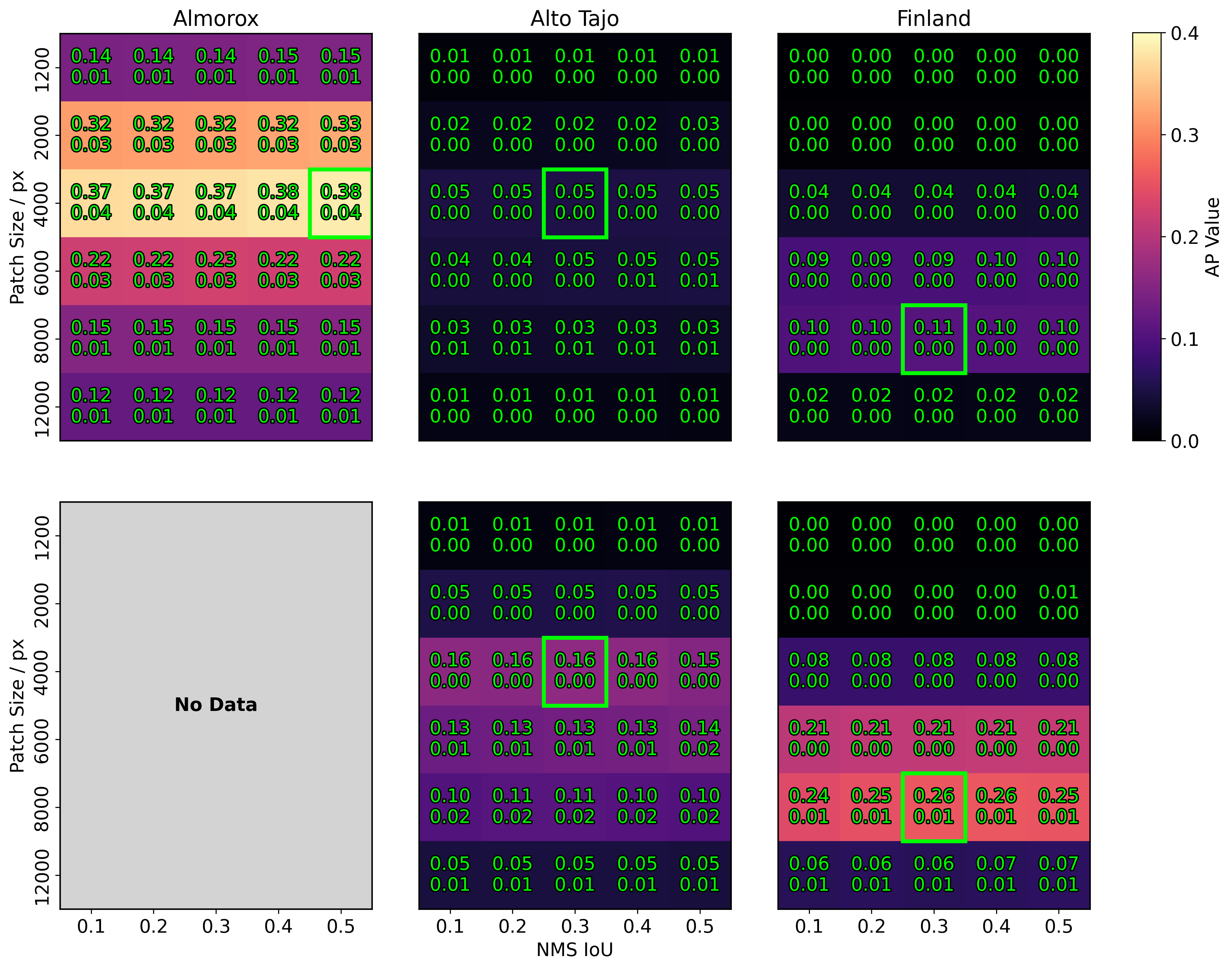}
\caption{Gridsearch results for \textbf{DeepForest} on Almorox (Manual labelling), Alto Tajo (TLS Labelling) and Joensuu (TLS Labelling). \textbf{(Top row)} Overall performance \textbf{(Bottom row)} Canopy performance. Top value within each cell denotes AP measured at an IoU of 0.5, and bottom values at 0.75. Cells coloured by AP50 for both overall and canopy, and best result outlined in green.}\label{fig:gsdeep}
\end{figure}

\begin{figure}
\centering
\includegraphics[width=15cm]{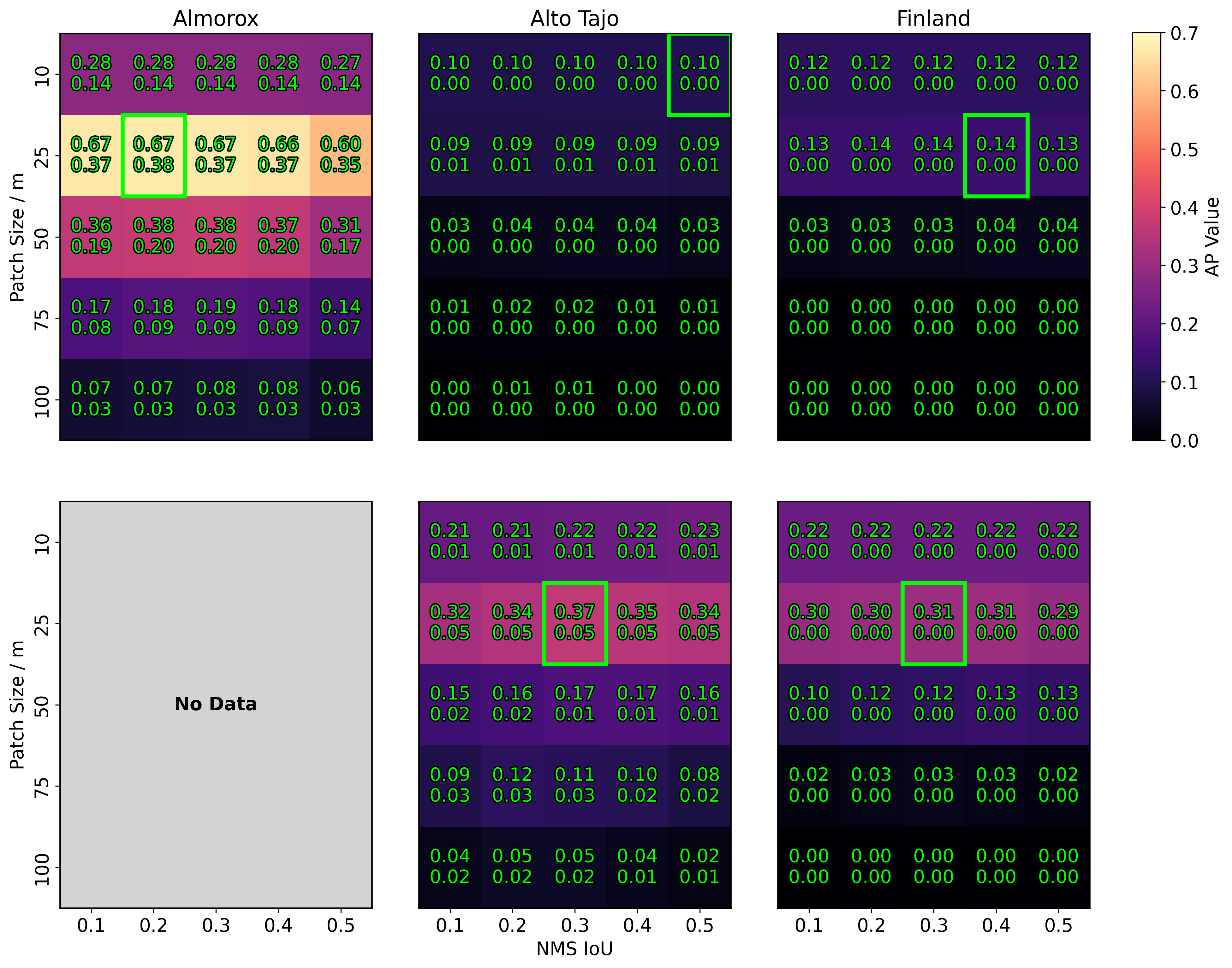}
\caption{Gridsearch results for \textbf{Detectree} on Almorox (Manual labelling), Alto Tajo (TLS Labelling) and Joensuu (TLS Labelling). \textbf{(Top row)} Overall performance. \textbf{(Bottom row)} Canopy performance. Top value within each cell denotes AP measured at an IoU of 0.5, and bottom values at 0.75. Cells coloured by AP50 for both overall and canopy, and best result outlined in green.}\label{fig:gsdet}
\end{figure}

Full results for the gridsearch outlined in Section~\ref{sec:modeleval} can be seen in Figure~\ref{fig:gsdeep} for DeepForest and Figure~\ref{fig:gsdet} for Detectree2. Note that, for Detectree2 on Alto Tajo (Middle column, Figure~\ref{fig:gsdet}), the optimum tile size was \SI{10}{\meter} despite our use of \SI{25}{\meter} in the main text. This was not coincident with the optimum tile size for canopy trees, unlike both models on each of the other datasets. This is likely because some canopy trees will not fit within a single \SI{10}{\meter} tile. We therefore chose to discard the results on \SI{10}{\meter} tiles in the main text, and use \SI{25}{\meter} tiles, which otherwise performed the best. Regardless, the overall performance of Detectree2 on the Alto Tajo data was near-constant when comparing \SI{10}{\meter} to \SI{25}{\meter} tiles (Figure~\ref{fig:gsdet}, middle column, top row) regardless, so the effect of this decision is likely insignificant.

\FloatBarrier
\section{Plot statistics}
\label{app:dataset-summary}
\begin{table}
\fontsize{10pt}{10pt}\selectfont
  \centering
  \begin{tabular}{llllll}
    \toprule
    \textbf{Site} & \textbf{Plot} & \textbf{Total Crowns} & \textbf{Max/Avg.} & \textbf{Orthomosaic Dim.} & \textbf{GSD} \\
    & & & \textbf{Crown Area} & &  \\
    & & & m$^2$ & Px & cmPx$^{-1}$ \\
    \midrule
    Finland & FIN01 & 47  & 30.6 / 14.0 & 13893$\times$19153 & 1.2 \\
    Finland & FIN02 & 59  & 28.8 / 11.1 & 14099$\times$16256 & 1.2 \\
    Finland & FIN04 & 42  & 34.7 / 15.1 & 13495$\times$19262 & 1.2 \\
    Finland & FIN09 & 123 & 27.0 / 7.5  & 12967$\times$18781 & 1.1 \\
    Finland & FIN10 & 72  & 27.9 / 12.6 & 13683$\times$20441 & 1.0 \\
    Finland & FIN12 & 141 & 29.7 / 5.2  & 14064$\times$18253 & 1.2 \\
    Finland & FIN13 & 112 & 21.2 / 6.2  & 14007$\times$16253 & 1.4 \\
    Finland & FIN15 & 90  & 25.6 / 8.6  & 11041$\times$14511 & 1.4 \\
    Finland & FIN18 & 84  & 25.9 / 9.6  & 15928$\times$18105 & 1.0 \\
    Finland & FIN19 & 67  & 25.3 / 12.2 & 15478$\times$21625 & 1.0 \\
    Finland & FIN21 & 56  & 32.0 / 12.5 & 10118$\times$13600 & 1.5 \\
    Finland & FIN23 & 101 & 16.9 / 7.8  & 12657$\times$17100 & 1.3 \\
    Finland & FIN25 & 98  & 26.6 / 9.0  & 13260$\times$18479 & 1.2 \\
    Finland & FIN27 & 78  & 20.0 / 8.5  & 13921$\times$16942 & 1.2 \\
    Finland & FIN28 & 51  & 30.2 / 13.5 & 13767$\times$17827 & 1.2 \\
    \midrule
    \textbf{Finland} & \textbf{Total/Avg.} & \textbf{1271} & \textbf{27.0 / 10.0} &  & \textbf{1.2} \\
    \midrule
    Spain   & SPA01 & 86  & 112.6 / 10.3 & 15815$\times$14739 & 1.6 \\
    Spain   & SPA02 & 39  & 24.1  / 10.4 & 13868$\times$16802 & 1.5 \\
    Spain   & SPA03 & 36  & 32.3  / 13.8 & 13240$\times$16325 & 1.4 \\
    Spain   & SPA05 & 61  & 116.8 / 10.8 & 16892$\times$13702 & 1.3 \\
    Spain   & SPA06 & 41  & 49.5  / 15.8 & 18162$\times$21429 & 1.1 \\
    Spain   & SPA13 & 79  & 48.2  / 7.3  & 14357$\times$17166 & 1.5 \\
    Spain   & SPA16 & 70  & 43.7  / 7.6  & 14481$\times$15887 & 1.4 \\
    Spain   & SPA17 & 125 & 40.5  / 7.4  & 18086$\times$22048 & 1.2 \\
    Spain   & SPA19 & 49  & 43.7  / 13.9 & 17561$\times$14619 & 1.2 \\
    Spain   & SPA23 & 35  & 36.3  / 9.8  & 13213$\times$17005 & 1.4 \\
    Spain   & SPA24 & 29  & 76.7  / 15.4 & 18613$\times$16119 & 1.3 \\
    Spain   & SPA25 & 51  & 58.1  / 13.7 & 14148$\times$16868 & 1.2 \\
    Spain   & SPA26 & 58  & 56.1  / 10.1 & 16343$\times$16965 & 1.4 \\
    Spain   & SPA27 & 22  & 61.2  / 16.3 & 17149$\times$15086 & 1.3 \\
    Spain   & SPA28 & 186 & 10.5  / 3.4  & 17314$\times$13230 & 1.6 \\
    Spain   & SPA29 & 121 & 85.7  / 6.7  & 16598$\times$17239 & 1.3 \\
    Spain   & SPA32 & 21  & 63.7  / 17.5 & 14276$\times$17126 & 1.6 \\
    Spain   & SPA35 & 41  & 43.6  / 9.3  & 16529$\times$19296 & 1.7 \\
    Spain   & SPA36 & 37  & 70.3  / 14.2 & 19592$\times$17082 & 1.7 \\
    \midrule
    \textbf{Spain} & \textbf{Total/Avg.} & \textbf{1387} & \textbf{52.6 / 10.7} &  & \textbf{1.4} \\
    \midrule
    \textbf{Combined} & \textbf{Total/Avg.} & \textbf{2658} & \textbf{40.3 / 10.4} &  & \textbf{1.3} \\
    \bottomrule
  \end{tabular}
\caption{\textbf{Summary statistics} for the 34 areas, across the two TLS-labelled sites, in our data. One plot is found in each area and the plot name is used to reference both the area (the entire area covered by the drone imagery) and the plot (a subset of the area covered by drone imagery that was also covered by TLS data). A total of 2658 crowns were deemed to be visible from above using the TLS data and thus segmented within the images.\label{tab:dataset-summary}}
\end{table}
\restoregeometry

\FloatBarrier
\section{Segmenting Orthomosaic Data Using TLS}

\subsection{Segmenting Orthoimagery using Point Clouds}
\label{app:seg}
In this work, we undertook the task of segmenting tree crowns as-seen from above using TLS point clouds collected from the ground. We segmented visible crown area only, and therefore sought to remove overlap between crowns in our dataset, such that our data contained only areas of crown clearly visible in aerial imagery. Methods tested on our data were therefore evaluated on their ability to segment crowns that are visible from above, which is not the same as total crown area, nor does it include all trees in the TLS dataset, since sub-canopy trees will not be visible. 

A simple approach to segment images might be to subtract crowns based on the maximum height of the respective tree. One advantage of this approach is that maximum height data is very often already available for existing crown segmentations. Such data could easily be processed into the desired form by simply assuming the crown of a taller tree is always above that of a shorter one, and subtracting the crown footprint of the taller from the smaller. This approach, however, does not consider height differences within crowns, which  may result in part of the crown of a shorter tree lying above that of a taller one. We developed a pipeline applicable to point cloud data to determine the tree corresponding to the highest layer of the canopy at all points across an area of interest.

Our pipeline begins with the assumption that individual trees have been fully segmented from whole-plot TLS point clouds. At the time of writing, the pipeline should be applicable to individual point clouds of any format supported by the Point Data Abstraction Library (PDAL, \cite{butler_pdal_2024}). A Digital Surface Model (DSM) was first created for each of the individual tree point clouds using PDAL, with a DSM resolution of 2cm (a typical resolution for downsampled TLS data, or an orthomosaic) and gap filling using a window size of 1 pixel (2cm). It is possible for a DSM with disconnected nonzero regions to form, if any regions of the point cloud can be separated by a lateral gap of more than the DSM resolution plus the window size. These parameters were tuned to minimise this effect while maintaining high visual fidelity for the resulting segmentation. This selection was based on visual interpretation. Users may need to select different parameters under different drone flight protocols or with different instruments. A DSM resolution roughly equal to the orthomosaic resolution and a window size roughly equal to TLS resolution would be a reasonable starting point. The minimum and maximum $x$ and $y$ coordinates (\texttt{minx}, \texttt{miny}, \texttt{maxx}, \texttt{maxy}) encountered across all trees were recorded at this stage, and updated for each new tree processed, resulting in the minimum and maximum $x$ and $y$ coordinates for the entire plot after all trees are processed.

Next, an empty GDAL Virtual Dataset (VRT) was created, spanning the entire plot (\texttt{minx}, \texttt{miny}, \texttt{maxx}, \texttt{maxy}), at the same per-pixel resolution as the individual DSMs. We read this VRT as a \texttt{numpy} \citep{harris_array_2020} array, \texttt{all\_array}, to enable pixel-wise operations. This array was constructed with two bands - one corresponding to a DSM of the entire plot, and one where the pixel value is equal to the index of the tallest tree at that location. We iterated through the DSMs for each tree one at a time, aligning them with the VRT of the entire site, reading them as \texttt{numpy} arrays and then updating both channels of \texttt{all\_array} using these individual arrays. After completing this loop, the layer containing the indices of the tallest tree at each pixel was polygonised using GDAL, resulting in a set of polygons where each polygon corresponds to a different tree. We then filtered these polygons, keeping only the largest polygon by area for each tree and removing any interior geometries.

This pipeline is shown as pseudocode in Algorithm~\ref{alg:tls_pipeline}. 

\section{TLS-Orthomosaic Alignment}
\label{app:ortho}
We made the following observations regarding generating orthomosaic data to align with TLS data that users may find useful should they need to apply this pipeline to new data.

\begin{itemize} 
    \item \textbf{Ground Control Point Accuracy}: Including GCPs measured with GNSS receivers in the field resulted in orthoimagery that did not align properly with TLS data. GCPs with lateral accuracy of 5-10 cm and vertical accuracy of 50-60 cm caused image misalignment during photogrammetric processing, leading to non-linear distortions in the orthomosaics. Since TLS data is precise at the level of individual branches, these distortions made alignment with orthomosaics impossible. Some software can adjust the alignment process based on GCP accuracy, which could help address these errors, although this in turn relies on accurate estimates of measurement uncertainty. Although Metashape 2.1.1 includes this functionality, we were unable to use it successfully despite having RMSE estimates for the GNSS coordinates. It is possible the RMSE estimates were also inaccurate. In contrast, generating orthoimagery without GCPs produced more accurate geometries, though the overall georeferencing accuracy was lower. In such cases, the entire image may be offset by several metres, but the geometries of individual crowns aligned better with the TLS point clouds. This type of error is straightforward to correct by hand using common GIS software, although potentially time consuming.
    \item \textbf{Photogrammetric Alignment}: Crown geometries were improved by not downsampling during alignment. 
    \item \textbf{Ground Plane Accuracy}: Accurate estimation of ground plane orientation (z-axis direction) is critical for proper data alignment. Even small errors of $3-4\degree$ in the z-axis direction caused distortions in the orthoimagery - as the resulting view orientation will not be vertical - making alignment with TLS data impossible. These distortions may not be immediately obvious without directly comparing the two data sources. Inaccurate altitude estimates for GCPs can introduce this type of error regardless of lateral accuracy. The use of accurately located common targets (Maximum RMSE 1-2cm) or GCPs in both the drone imagery and TLS scans is likely the best approach to achieving strong aligment across the two data sources - noting that both their vertical and lateral accuracies must both meet this requirement.
    \item \textbf{Orthomosaic Artefacting}: Previous work constructing orthomosaic data of forests~\citep{troles_bamforests_2024} for aerial segmentation reported small artefacts in the resulting images. We found that constructing a 3D model (corresponding to a surface type set to `Arbitrary' in Metashape) from Depth Map data and then building the orthomosaics on the 3D model eliminated most of these artefacts compared to building orthomosaics on Digital Elevation Model (DEM) data. 
\end{itemize}